\documentclass[runningheads]{llncs}
\usepackage{graphicx}

\usepackage{tikz}
\usepackage{comment} 
\usepackage{amsmath,amssymb} 
\usepackage{color}

\usepackage{graphicx}
\usepackage{comment}
\usepackage{amsmath,amssymb} 
\usepackage{color}
\usepackage[utf8]{inputenc} 
\usepackage[T1]{fontenc}    
\usepackage{hyperref}       
\usepackage{url}            
\usepackage{booktabs}       
\usepackage{amsfonts}       
\usepackage{amsmath}
\usepackage{amssymb}
\usepackage{nicefrac}       
\usepackage{microtype}      
\usepackage{adjustbox}
\usepackage{booktabs}
\usepackage{multirow}
\usepackage{adjustbox}
\usepackage{xcolor}
\usepackage{graphics,psfrag}
\usepackage{diagbox} 
\usepackage{subfigure}
\usepackage{bbding}
\usepackage{amssymb}
\usepackage{wrapfig}
\usepackage{bbding}

\def\ie{\emph{i.e.}}

\def\etal{\emph{et al.}}

\begin{document}
\pagestyle{headings}
\mainmatter
\def\ECCVSubNumber{1698}  

\title{Video Super-Resolution with Recurrent Structure-Detail Network} 

%

\titlerunning{Video Super-Resolution with Recurrent Structure-Detail Network}
%


\author{Takashi Isobe\inst{1,2}\thanks{The work was done in Noah's Ark Lab, Huawei Technologies.} \and
	 Xu Jia\inst{2}\textsuperscript{\Envelope} \and
	Shuhang Gu\inst{3} \and
	Songjiang Li\inst{2} \and
	Shengjin Wang \inst{1}\textsuperscript{\Envelope} \and \\
	Qi Tian \inst{2} 
}

\authorrunning{T. Isobe et al.}
%
\institute{Department of Electronic Engineering, Tsinghua University
\email{	{\texttt{\small{jbj18@mails.tsinghua.edu.cn}}}, {\texttt{\small{wgsgj@tsinghua.edu.cn}}}}\\\and
Noah's Ark Lab, Huawei Technologies \\
\email{\texttt{\small{\{x.jia, songjiang.li, tian.qi1\}@huawei.com}}}\\\and
School of Eie, The University of Sydney \\
\email{shuhanggu@gmail.com}
}


\maketitle

\begin{abstract}
	Most video super-resolution methods super-resolve a single reference frame with the help of neighboring frames in a temporal sliding window. They are less efficient compared to the recurrent-based methods. In this work, we propose a novel recurrent video super-resolution method which is both effective and efficient in exploiting previous frames to super-resolve the current frame. It divides the input into structure and detail components which are fed to a recurrent unit composed of several proposed two-stream structure-detail blocks. In addition, a hidden state adaptation module that allows the current frame to selectively use information from hidden state is introduced to enhance its robustness to appearance change and error accumulation. Extensive ablation study validate the effectiveness of the proposed modules. Experiments on several benchmark datasets demonstrate superior performance of the proposed method compared to state-of-the-art methods on video super-resolution. Code is available at \url{https://github.com/junpan19/RSDN}.
	
\end{abstract}
\section{Introduction}
\label{intro}

Super-resolution is one of the fundamental problem in image processing, which aims at reconstructing a high resolution (HR) image from a single low-resolution (LR) image or a sequence of LR images. 
According to the number of input frames, the field of SR can be divided into two categories,~\ie, single image super-resolution (SISR) and multi-frame super-resolution (MFSR). 
For SISR, the key issue is to exploit natural image prior for compensating missing details; while for MFSR, how to take full advantage from additional temporal information is of pivotal importance. 
In this work, we focus on the video super-resolution (VSR) task which belongs to MFSR. It draws much attention in both research and industrial communities because of its great value on computational photography and surveillance.

In the last several years, great attempts have been made to exploit multi-frame information for VSR.
One category of approaches utilize multi-frame information by conducting explicit motion compensation.
These approaches \cite{caballero2017real,kappeler2016video,xue2019video,tao2017detail,liu2017robust} firstly compute optical flow between a reference frame and neighboring frames and then employ the aligned observations to reconstruct the high-resolution reference frame.
However, estimating dense optical flow itself is a challenging and time-consuming task.
Inaccurate flow estimation often leads to unsatisfactory artifacts in the SR results of these flow-based VSR approaches.
In addition, the heavy computational burden also impedes the application of these applications in resource-constrained devices and time-sensitive scenarios.
In order to avoid explicit motion compensation, another category of methods propose to exploit the motion information in an implicit manner.
The dynamic upsampling filters \cite{jo2018deep} and the progressive fusion residual blocks \cite{yi2019progressive} are designed to explore flow-free motion compensation.
With these flexible compensation strategies, \cite{jo2018deep,yi2019progressive} not only avoid heavy motion estimation step but also achieve highly competitive VSR performance.
However, they still suffer from the redundant computation for several neighboring frames within a temporal window and need to cache several frames in advance to conduct VSR.
Recently, for the pursuit of efficiency, there is an emerging trend of applying recurrent connection to address the VSR task.
These approaches~\cite{sajjadi2018frame,Fuoli-arxiv19-rlsp} make use of recurrent connection to conduct video super-resolution in a streaming way, that is, output or hidden state of previous time steps is used to help super-resolve future frames. In addition, they are able to exploit temporal information from many frames in the past. By simply propagating output and hidden state of previous steps with a recurrent unit, they achieve promising VSR performance with considerably less processing time.

\begin{figure}[t]
	\centering
	\includegraphics[width=\textwidth]{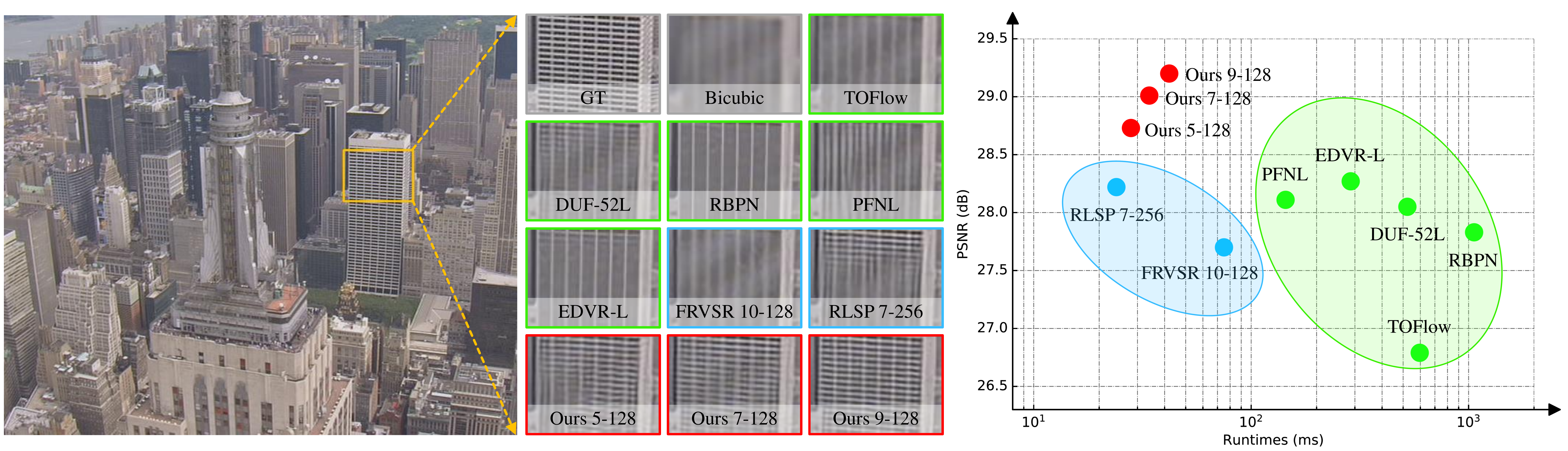}
	\caption{VSR results on the \textbf{City} sequence in Vid4. Our method produces finer details and stronger edges with better balance between speed and performance than both temporal sliding window based~\cite{xue2019video,jo2018deep,haris2019recurrent,wang2019edvr,yi2019progressive} and recurrent based methods~\cite{sajjadi2018frame,Fuoli-arxiv19-rlsp}. {\color{blue}Blue} box represents recurrent-based and {\color{green}green} box represents sliding window based methods. Runtimes (ms) are calculated on an HR image of size 704$\times$576. }
	\label{fig:runtime}
 	\vspace{-5mm}
\end{figure}

In this paper, we propose a novel recurrent network for efficient and effective video super-resolution.
Instead of simply concatenating consecutive three frames with previous hidden state as in~\cite{Fuoli-arxiv19-rlsp}, we propose to decompose each frame of a sequence into components of structure and detail and aggregate both current and previous structure and detail information to super-resolve each frame. 
Such a strategy not only allows our method to address different difficulties in the structure and detail components,
but also able to impose flexible supervision to recover high-frequency details and strengthen edges in the reconstruction.

In addition, we observe that hidden state in a recurrent network captures different typical appearances of a scene over time. To make full use of temporal information in hidden state, we treat the hidden state as a historical dictionary and compute correlation between the reference frame and each channel in hidden state. This allows the current frame to highlight the potentially helpful information and suppress outdated information such that information fusion would be more robust to appearance change and accumulated errors.
Extensive ablation study demonstrates the effectiveness of the proposed method. It performs very favorably against state-of-the-art methods on several benchmark datasets, in both super-resolution performance and speed.

\section{Related Work}
\label{related}

\subsection{Single Image Super-Resolution}
Traditional SISR methods include interpolation-based methods and dictionary learning-based methods. However, since the rise of deep learning, most traditional methods are outperformed by deep learning based methods. A simple three-layer CNN is proposed by Dong~\cite{dong2014learning}, showing great potential of deep learning in super-resolution for the first time. Since then, plenty of new network architectures~\cite{kim2016accurate,ledig2017photo,lim2017enhanced,zhang2018residual,haris2018deep,zhang2018image} have been designed to explore power of deep learning to further improve performance of SISR. In addition, researchers also investigate the role of losses for better perceptual quality. More discussions can be found in a recent survey~\cite{yang2019deep}. A very relevant work is the DualCNN method proposed by Pan~\etal~\cite{pan2018learning}, where authors proposed a network with two parallel branches to reconstruct structure and detail components of an image, respectively. However, different from that work, our method aims at addressing the video super-resolution task. It decomposes the input frames into structure and detail components and propagates them with a recurrent unit that is composed of two interleaved branches to reconstruct the high-resolution targets. It is motivated by the assumption that structure and detail components not only suffer from different difficulties in high-resolution reconstruction but also take benefit from other frames in different ways.

\subsection{Video Super-Resolution}
Although SISR methods can also be used to address the video super-resolution task, they are not very effective because they only learn to explore natural prior and self-similarity within an image and ignore rich temporal information in a sequence. The key to video super-resolution is to make full use of complementary information across frames. Most video super-resolution methods can be roughly divided into two categories according to whether they conduct motion compensation in an explicit way or not.

{\flushleft{\textbf{Explicit motion compensation.}}}
 Most methods with explicit motion compensation follow a pipeline of motion estimation, motion compensation, information fusion and upsampling. VESPCN~\cite{caballero2017real} presents a joint motion compensation and video super-resolution with a coarse-to-fine spatial transformer module. Tao~\etal~\cite{tao2017detail} proposed an SPMC module for sub-pixel motion compensation and used a ConvLSTM to fuse information across aligned frames. Xue~\etal~\cite{xue2019video} proposed a task-oriented flow module that is trained together with a video processing network for video denoising, deblock or super-resolution. 
 In~\cite{sajjadi2018frame}, Sajjadi~\etal~proposed to super-resolve a sequence of frames in a recurrent manner, where the result of previous frame is warped to the current frame and two frames are concatenated for video super-resolution.
 Haris~\etal~\cite{haris2019recurrent} proposed to use a recurrent encoder-decoder module to exploit explicitly estimated inter-frame motion.
 Wang~\etal~\cite{wang2019edvr} proposed to align multiple frames to a reference frame in feature space with a deformable convolution based module and fuse aligned frames with a temporal and spatial attention module.
 However, the major drawback of such methods is the heavy computational load introduced by motion estimation and motion compensation.

{\flushleft{\textbf{Implicit motion compensation.}}} 
As for methods with implicit motion compensation~\cite{jo2018deep,yi2019progressive,haris2019recurrent,Fuoli-arxiv19-rlsp}, they do not estimate motion between frames and align them to a reference frame, but focus on designing an advanced fusion module such that it can make full use of complementary information across frames.
Jo~\etal~\cite{jo2018deep}~proposed to use a 3D CNN to exploit spatial-temporal information and predict a dynamic upsampling filter to reconstruct HR images.
In~\cite{yi2019progressive}, Yi~\etal~proposed to fuse spatial-temporal information across frames in a progressive way and use a non-local module to avoid explicit motion compensation.
Video super-resolution with implicit motion can also be done with recurrent connection. 
Huang~\etal~\cite{huang2015bidirectional} proposed a bidirectional recurrent convolutional network to model temporal information across multiple frames for efficient video super-resolution.
In~\cite{Fuoli-arxiv19-rlsp}, Fuoli~\etal~proposed to conduct temporal information propagation with a recurrent architecture in feature space.
Our method also adopts the recurrent way to conduct video super-resolution without explicit motion compensation. However, different from the above methods, we proposed to decompose a frame into two components of structure and detail and propagate them separately. In addition, we also compute correlation between the current frame and the hidden state to adaptively use the history information in the hidden state for better performance and less
risk of error accumulation.

\subsection{Recurrent Networks for Video-based Tasks}
Recurrent networks have been widely used in different video recognition tasks. Donahue~\etal~\cite{kim2016deeply} proposed a class of recurrent convolutional architectures which combine convolutional layers and long-range temporal information for action recognition and image captioning.
In~\cite{singh2016multi}, a bi-directional LSTM is applied after a multi-stream CNN to fully explore temporal information in a sequence for fine-grained action detection.
Du~\etal~\cite{du2017rpan} proposed a recurrent network with a pose attention mechanism which exploits spatial-temporal evolution of human pose to assist action recognition.
Recurrent networks are capable of processing sequential information by integrating information from each frame in their hidden states. 
They can not only be used for high-level video recognition tasks but are also suitable for low-level video processing tasks.
\begin{figure}[t]
	\centering
	\includegraphics[width=\textwidth]{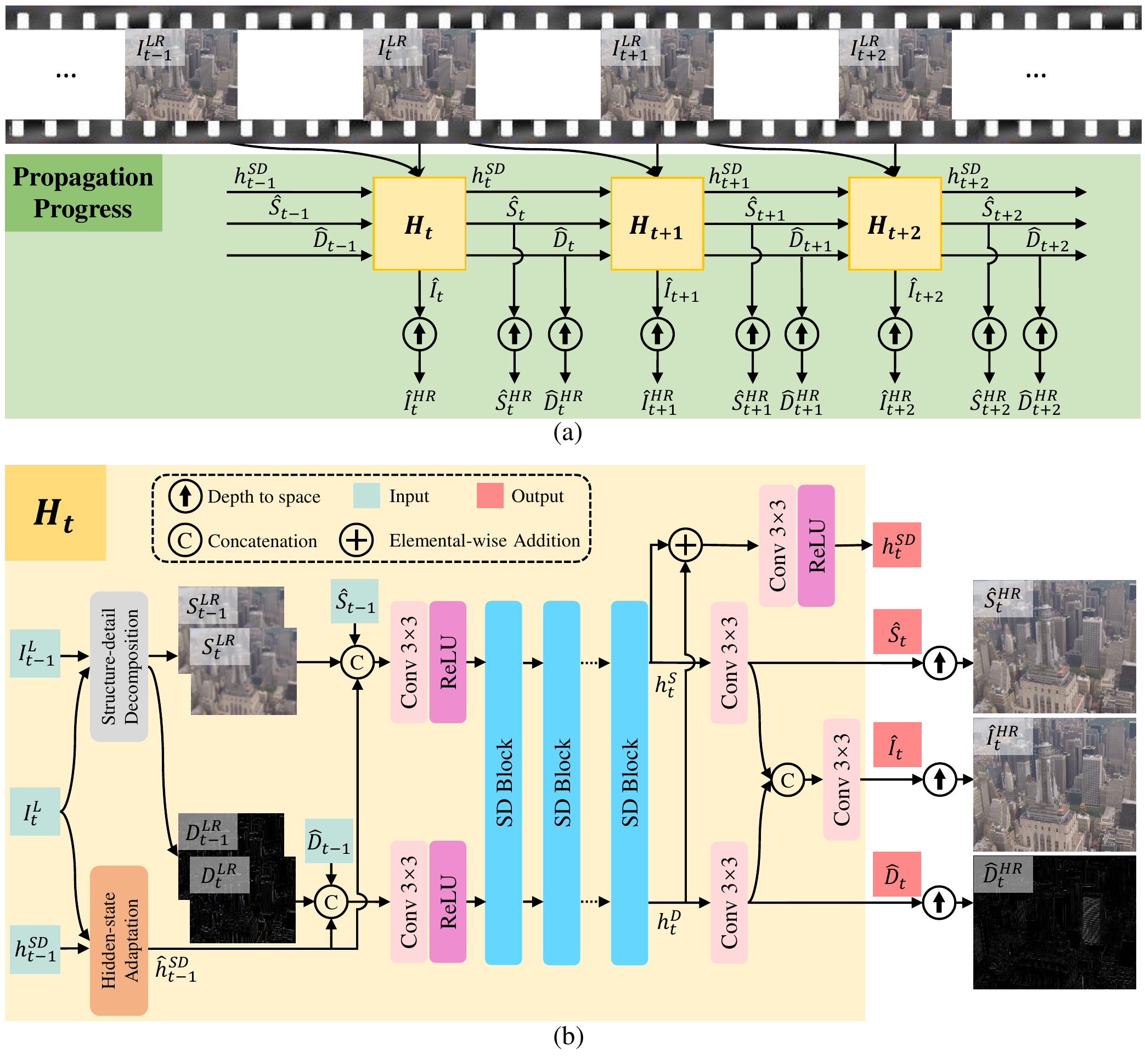}
	\caption{(a) The overall pipeline of the proposed method; (b) architecture of the recurrent structure-detail unit.}
	\label{fig:pipeline}
	\vspace{-5mm}
\end{figure}
\section{Method}
\label{method}
\subsection{Overview}

Given a low-resolution video clip $\{I^{LR}_{1:N}\}$, $N\ge 2$,
the goal of VSR is to produce a high-resolution video sequence $\{\hat{I}^{HR}_{1:N}\}$ from the corresponding low-resolution one by filling in missing details for each frame. In order to process a sequence efficiently, we conduct VSR in a recurrent way similar to~\cite{sajjadi2018frame,Fuoli-arxiv19-rlsp}. However, instead of feeding a whole frame to a recurrent network at each time step, we decompose each input frame into two components,~\ie, a structure component and a detail component, to the following network. Two kinds of information interact with each other in the proposed SD blocks over time, which is not only able to sharpen the structure of each frame but also manages to recovers missing details. In addition, to make full use of complementary information stored in hidden states, we treat hidden state as a history dictionary and adapt this dictionary to the demand of the current frame. This allow us to highlight the potential helpful information and suppress outdated information. The overall pipeline is shown in Fig.~\ref{fig:pipeline}(a).

\begin{figure}[t]
	\centering
	\includegraphics[width=\textwidth]{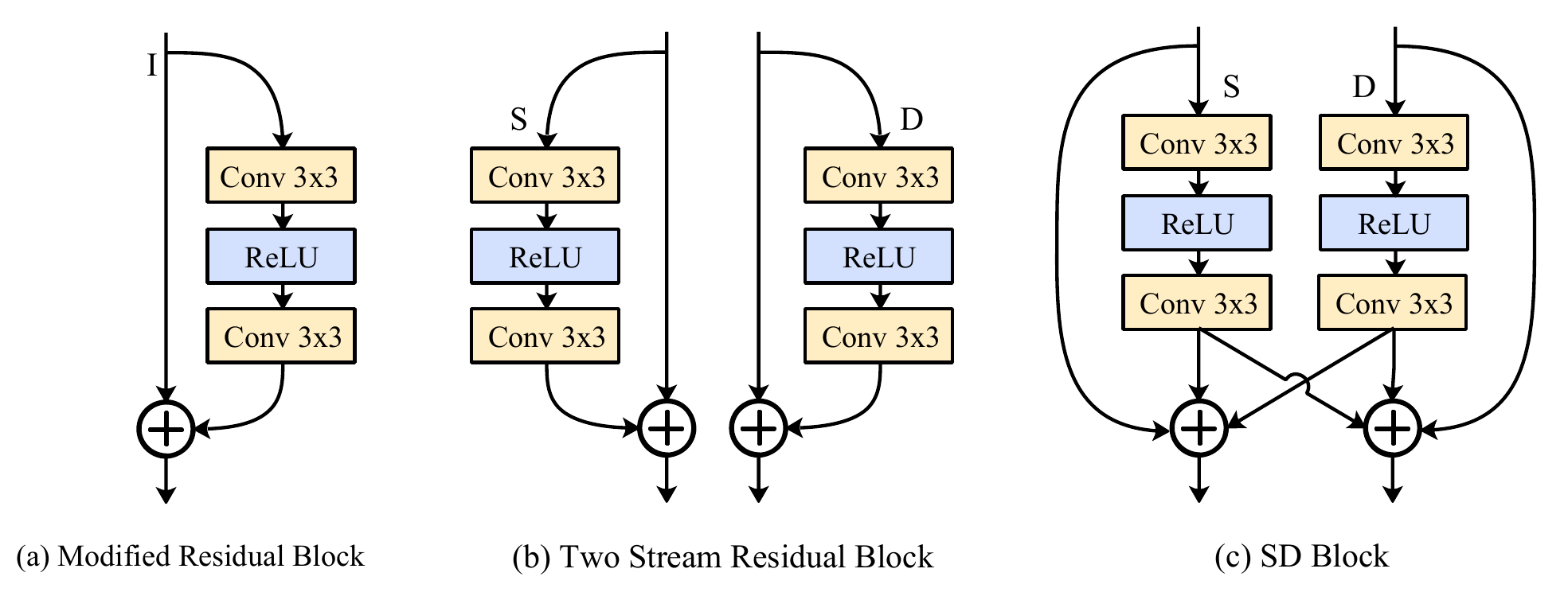}
 	\vspace{-3mm}
	\caption{Variant design for Structure-Detail block.}
	\label{fig:SD_block}
	\vspace{-5mm}
\end{figure}

\subsection{Recurrent Structure-Detail Network}
{\flushleft{\textbf{Recurrent unit.}}}
Each frame can be decomposed into a structure component and a detail component. The structure component models low-frequency information in an image and motion between frames. While the detail component captures fine high-frequency information and slight change in appearance. These two components suffer from different difficulty in high-resolution reconstruction and take different benefit from other frames, hence should be processed separately.

In this work, we simply apply a pair of bicubic downsampling and upsampling operations to extract structural information from a frame $I^{LR}_t$, which is denoted as $S^{LR}_t$. The detail component $D^{LR}_t$ can be then computed as the difference between the input frame $I^{LR}_t$ and the structure component $S^{LR}_t$. In fact, we can also use other ways such as low-pass filtering and high-pass filtering to get these two components. 
For simplicity, we adopt a symmetric architecture for the two components in the recurrent unit, as shown in Fig.~\ref{fig:pipeline} (b). 
Here we only take D-branch at time step $t$ as an example to explain its architecture design. 
Detail components of the previous and current frames $\{D^{LR}_{t-1},D^{LR}_t\}$ are concatenated with the previously estimated detail map $\hat{D}_{t-1}$ and hidden state $\hat{h}^{SD}_{t-1}$ along the channel axis. Such information is further fused by one $3\times3$ convolutional layer and several structure-detail (SD) blocks. In this way, this recurrent unit manages to integrate together information from two consecutive input frames, output of the previous time step and historical information stored in the hidden state. $h^D_t$ denotes the feature computed after several SD blocks. It goes through another $3 \times 3$ convolutional layer and an upsampling layer to produce the high resolution detail component $\hat{D}^{HR}_t$. The S-branch is designed in a similar way. $h^S_t$ and $h^D_t$ are combined to produce the final high resolution image $\hat{I}^{HR}_t$ and new hidden state $h^{SD}_t$.
The D-branch focuses on extracting complementary details from past frames for the current frame while the S-branch focuses on enhancing existed edges and textures in the current frame.

{\flushleft{\textbf{Structure-Detail block.}}}
Residual block~\cite{ledig2017photo} and dense block~\cite{huang2017densely} are widely used in both high-level and low-level computer vision tasks because of their effectiveness in mining and propagating information. 
In this section, we compare several variants of blocks in propagating information in a recurrent structure-detail unit. 
For comparison, we also include a modified residual block as shown in Fig.~\ref{fig:SD_block}(a), which only has one branch and takes the whole frames as input. To adapt it to address two branches, the easiest way is to have two modified residual blocks that process two branches separately, as shown in Fig.~\ref{fig:SD_block}(b). However, in this way each branch only sees the component-specific information and can not makes full use of all information in the input frames. Therefore, we propose a new module called structure-detail (SD) block, as shown in Fig.~\ref{fig:SD_block}(c). The two components are first fed to two individual branches and then combined with an addition operation.
In this way, it not only specializes on each component but also promotes information exchange between structure and detail components. Its advantage over the other two variants is validated in the experiment section.

\begin{wrapfigure}{R}{0.5\textwidth}
  \vspace{-30pt}
  \begin{center}
    \includegraphics[width=0.5\textwidth]{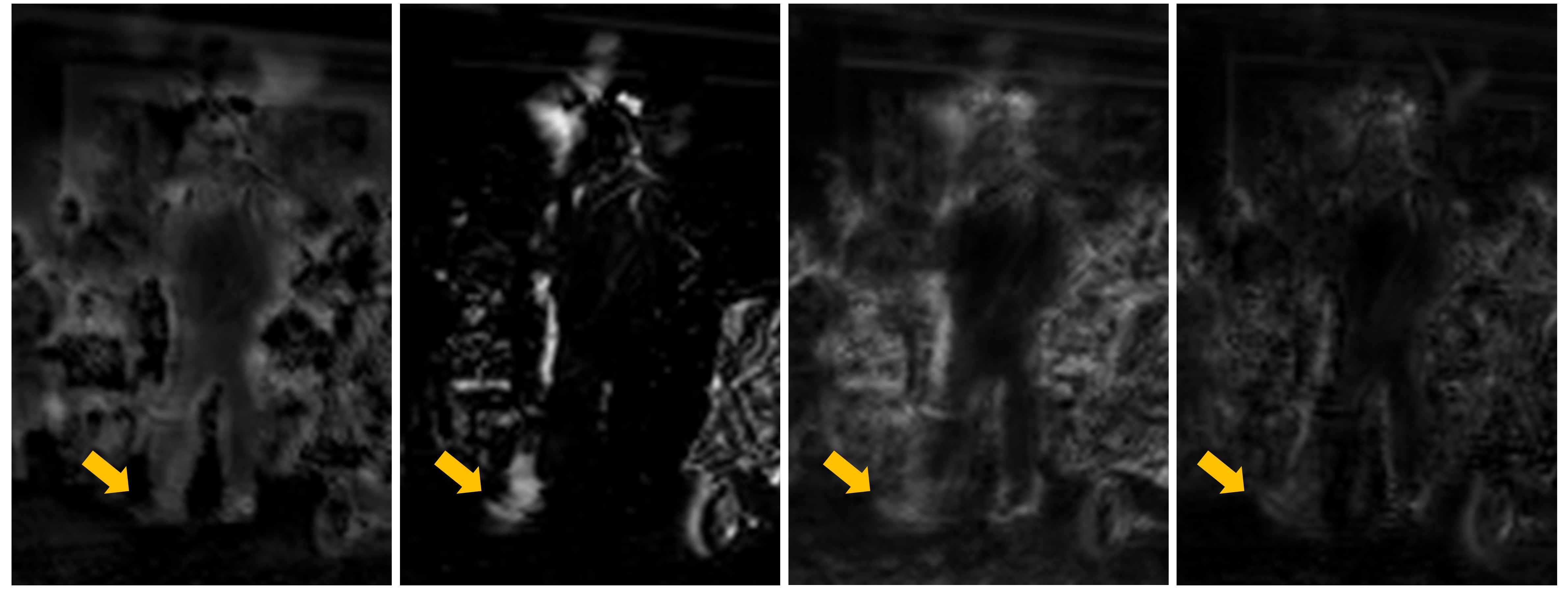}	
  \end{center}  
  \vspace{-5mm}
   \caption{Four channels in hidden state at a certain time step are selected for visualization. Yellow arrow denotes the difference in appearance among these four channels. Zoom in for better visualization.}
  \label{fig:hidden_state}
  \vspace{-3mm}
\end{wrapfigure}

\subsection{Hidden State Adaptation}
In a recurrent neural network, hidden state at time step $t$ would summarize past information in the previous frames. When applying a recurrent neural network to the video super-resolution task, hidden state is expected to model how a scene's appearance evolves over time, including both structure and detail. The previous recurrent-based VSR method~\cite{Fuoli-arxiv19-rlsp} directly concatenates previous hidden state and two input frames and feeds it to several convolutional layers. However, for each LR frame to be super-resolved, it has distinct appearance and is expected to borrow complementary information from previous frames in different ways. Applying the same integration manner to all frames is not optimal and could hurt the final performance. As shown in Fig.~\ref{fig:hidden_state}, different channels in hidden state describe different scene appeared in the past. They should make different contribution to different positions of different frames, especially when there are occlusion and large deformation with some channels of the hidden state. 

\begin{figure}[t]
	\centering
	\includegraphics[width=\textwidth]{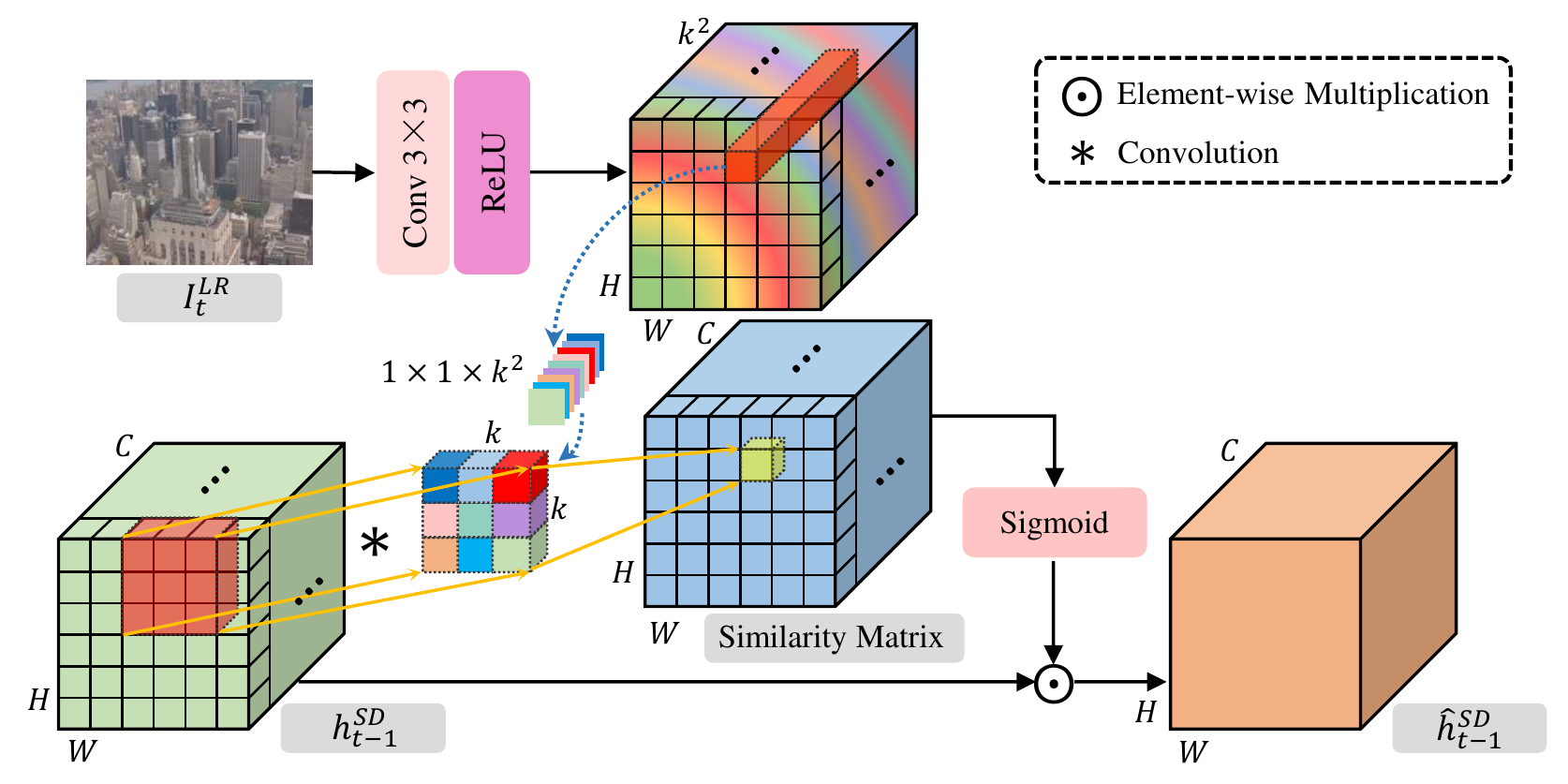}
	\caption{Design of hidden state adaptation module.}
	\label{fig:HSA}
	\vspace{-5mm}
\end{figure}
In this work, we proposed the Hidden State Adaptation (HSA) module to adapt a hidden state to the appearance of the current frame. As for each unit in hidden state, it should be highlighted if it has similar appearance as the current frame; otherwise, it should be suppressed if it looks very different.  
With this module the proposed method carefully chooses only useful information in previous frames, hence  alleviate the influence of drastic appearance change and error accumulation.
Since response of a filter models correlation between the filter and a neighborhood on an image, here we take similar way to compute correlation between an input frame and hidden state. 
Inspired by~\cite{jia2016dynamic}, we generate spatially variant and sample specific filters for each position in the current frame and use those filters to compute their correlation with the corresponding positions in each channel of hidden state.
Specifically, these spatially variant filters $F^\theta_t \in R^{ H \times W  \times (k \times k)}$ are obtained by feeding the current frame $I^{LR}_t \in R^{H\times W \times 3} $ into a convolutional layer with ReLU activation function~\cite{glorot2011deep}, where $H$ and $W$ are respectively height and width of the current frame, and $k$ denotes the size of filters. 
Then, each filter $F^\theta_t(i,j)$ are applied to a $k\times k$ window of $h^{SD}_{t-1}$ centered at position $(i,j)$ to conduct spatially variant filtering. This process can be formulated as: 
\begin{equation}
	M_{t}(i,j,c) = \sum_{u=-\lceil k/2 \rfloor}^{\lfloor k/2 \rfloor} \sum_{v=-\lfloor k/2 \rfloor}^{\lfloor k/2 \rfloor} F^\theta_t(i,j,u,v) \times 
	h^{SD}_{t-1}(i+u,j+v,c),
\end{equation}
where $M_{t}(i,j,c)$ represents correlation between the current frame and the $c$-th channel of hidden state at position $(i,j)$.

It is further fed to a sigmoid activation function $\sigma(\cdot)$ that transforms it into a similarity value in range $[0,1]$.
Finally, the adapted hidden state $\hat{h}^{SD}_{t-1}$ is computed by:
\begin{equation}
\hat{h}^{SD}_{t-1} = M_t \odot h^{SD}_{t-1},
\end{equation} 
where `$\odot$' denotes element-wise multiplication. 


\subsection{Loss functions}
Since the proposed recurrent network has two streams, the trade-off between supervision on structure and detail during training is very important. Imbalanced supervision on structure and detail might produce either sharpened frames but with less details or frames with many weak edges and details. Therefore, we propose to train the proposed network with three loss terms as shown in eq.~\ref{eq:total_loss}, one for structure component, one for detail component, and one for the whole frame. $\alpha$, $\beta$ and $\gamma$ are hyper-parameters to balance the trade-off of these three terms. The loss to train an $N$-frame sequence is formulated as:
\begin{equation}
\mathcal{L} = \dfrac{1}{N} \sum_{t=1}^{N}(\alpha \mathcal{L}^\mathcal{S}_t + \beta \mathcal{L}^\mathcal{D}_t + \gamma \mathcal{L}^\mathcal{I}_t).
\label{eq:total_loss}
\end{equation}
Similar to~\cite{lai2018fast}, we use Charbonnier loss function to compute the difference between reconstruction and high-resolution targets. Hence, we have $\mathcal{L}^\mathcal{S}_t = \sqrt{\Vert S^{HR}_t - \hat{S}^{HR}_t \Vert^2 + \varepsilon^2}$ for structure component, 
$\mathcal{L}^\mathcal{D}_t = \sqrt{\Vert D^{HR}_t - \hat{D}^{HR}_t \Vert^2 + \varepsilon^2}$ for detail component, 
and $\mathcal{L}^\mathcal{I}_t = \sqrt{\Vert I^{HR}_t - \hat{I}^{HR}_t \Vert^2 + \varepsilon^2}$ for the whole frame. The effectiveness of these loss functions is validated in the experiment section. 
\section{Experiments}
\label{experiment}
In this section, we first explain the experiment datasets and implementation details of the proposed method. Then extensive ablation study is conducted to analyze the effectiveness of the proposed SD block and hidden state adaptation module. Furthermore, the proposed method is compared with state-of-the-art video super-resolution methods in terms of both effectiveness and efficiency.

\subsection{Implementation Details}
\paragraph{\textbf{Datasets.}}
Some works~\cite{sajjadi2018frame,yi2019progressive} collect private training data from youtube on their own, which is not suitable for fair comparison with other methods. In this work, we adopt a widely used video processing dataset Vimeo-90K to train video super-resolution models. Vimeo-90K is a recent proposed large dataset for video processing tasks, which contains about $90K$ 7-frame video clips with various motions and diverse scenes. About $7K$ video clips select out of $90K$ as the test set, termed as Vimeo-90K-T. To train our model, we crop patches of size $256\times 256$ from HR video sequences as the target. Similar to~\cite{jo2018deep,sajjadi2018frame,Fuoli-arxiv19-rlsp,yi2019progressive}, the corresponding low-resolution patches are obtained by applying Gaussian blur with $\sigma=1.6$ to the target patches followed by $\times 4$ times downsampling.

To validate the effectiveness of the proposed method, we evaluate our models on several popular benchmark datasets, including Vimeo-90K-T~\cite{xue2019video}, Vid4~\cite{liu2013bayesian} and UDM10~\cite{yi2019progressive}. As mentioned above, Vimeo-90K-T contains a lot of video clips, but each clip has only 7 frames. Vid4 and UDM10 are long sequences with diverse scenes, which is suitable to evaluate the effectiveness of recurrent-based method in information accumulation~\cite{sajjadi2018frame,Fuoli-arxiv19-rlsp}.

{\flushleft{\textbf{Training Details.}}}
The base model of our method consists of 5 SD blocks where each convolutional layer has 128 channels,~\ie, RSDN 5-128. By adding more SD blocks, we can obtain RSDN 7-128 and RSDN 9-128. The performance can be further boosted with only small increase on computational cost and runtime. We adopt $K=3$ for HSA module for efficiency. 
To fully utilize all given frames, we pad each sequence by reflecting the second frame at the beginning of the sequence.
When dealing with the first frame of a sequence, the previous estimated detail $\hat{D}_{t-1}$, structure $\hat{S}_{t-1}$ and hidden state feature $h^{SD}_{t-1}$ are all initialized with zeros. 
The model training is supervised with Charbonnier penalty loss function and is optimized with Adam optimizer~\cite{kingma2014adam} with $\beta_1=0.9$ and $\beta_2=0.999$. 
Each mini-batch consists of $16$ samples. The learning rate is initially set to $1 \times 10^{-4}$ and is later down-scaled by a factor of 0.1 every $60$ epoch till $70$ epochs. The training data is augmented by standard flipping and rotating. All experiments are conducted on a server with Python 3.6.4, PyTorch 1.1 and Nvidia Tesla V100 GPU. 

\begin{table}[t]
	\centering
	\scalebox{0.67}{	
		\begin{tabular}{c||c|c||c|c|c||c|c|c}
			\hline
			\textbf{Method} & \multicolumn{2}{c||}{\begin{tabular}{@{}c@{}} One Stream\\7-256 \end{tabular}}   & \multicolumn{3}{c||}{\begin{tabular}{@{}c@{}}Two Stream\\7-128 \end{tabular}} & \multicolumn{3}{c}{\begin{tabular}{@{}c@{}} SD Block\\7-128 \end{tabular}}   \\ \hline \hline
			\textbf{Model} & Model 1 &Model 2   & Model 3  &Model 4  &Model 5  &Model 6  &Model 7  &Model 8  \\ 
			\textbf{HSA?} & w/o & w/ & w/  & w/o &w/  &w/ &w/o & w/ \\ 
			\textbf{Input}& Image &Image  &Image &S~\&~D &S~\&~D  &Image &S~\&~D  &S~\&~D 
			 \\ \hline
			PSNR/SSIM &27.58/0.8410  &27.65/0.8444   &27.70/0.8452 &27.64/0.8404  &27.68/0.8429   &27.73/0.8460  &27.76/0.8463 &{27.79}/{0.8474} \\ \hline
		\end{tabular}
	}
	\vspace{3mm}
	\caption{Ablation study on different network architecture.}
	\vspace{-5mm}
	\label{table:block}
\end{table}

{\flushleft{\textbf{Recurrent Unit.}}}
We compare three kinds of blocks for information flow in the recurrent unit,~\ie, the three blocks shown in Fig.~\ref{fig:SD_block}. 
For fair comparison among these blocks, we keep these three networks with almost the same parameters by setting the channel of convolutional layers in model 1 to 256, and setting the one in model 4 and 7 to 128.
\subsection{Ablation Study}
In this section, we conduct several ablation studies to analyze the effectiveness of the proposed SD block and the hidden state adaptation module. In addition, we also investigate the influence of different supervision on structure and detail components on the reconstruction performance.
\begin{wrapfigure}{R}{0.6\textwidth}
  \vspace{-30pt}
  \begin{center}
    \includegraphics[width=0.6\textwidth]{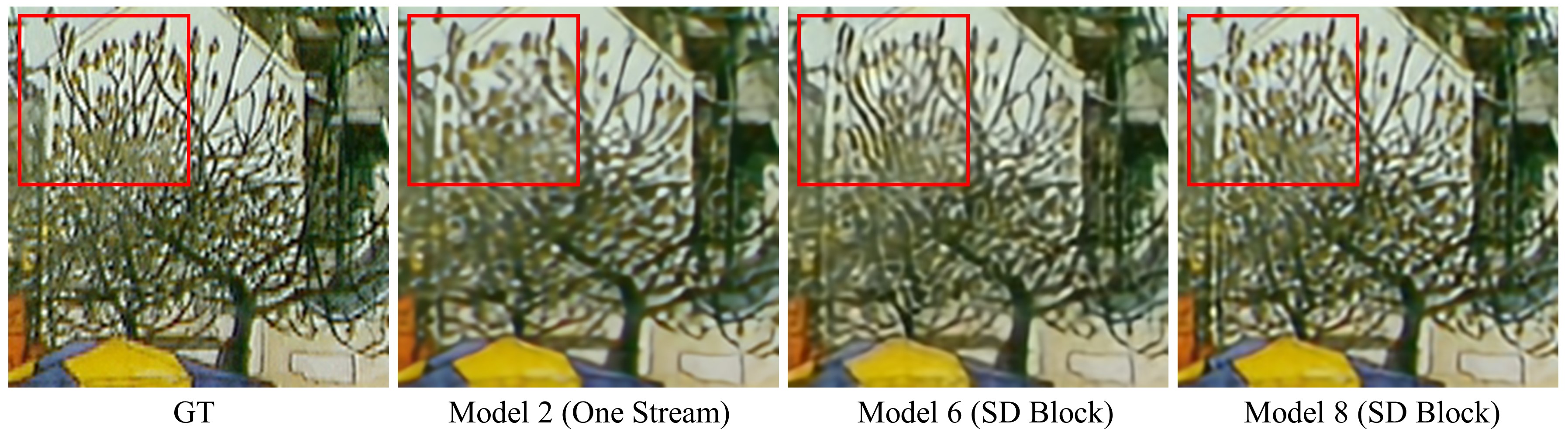}	
  \end{center}  
  \vspace{-5mm}
 \caption{ Qualitative comparison between different network structures. Zoom in to see better visualization.}
  \label{fig:model_ablated}
  \vspace{-3mm}
\end{wrapfigure}
As shown in Tab.~\ref{table:block}, model 1 and model 4 achieves similar performance, with model 1 a little higher SSIM and model 4 a little higher PSNR. This implies that simply dividing the input into structure and detail components and processing each one individually does not work well. Although it seems that having two branches to process each component divides a difficult task into two easier ones, it makes each one blind to the other and can not make full use of the information in the input to reconstruct either component.

By introducing information exchange between structure and detail components, model 7 obtains better performance than model 1 and 4 in both PSNR and SSIM. Similar result can also found in comparison among model 2, 5 and 8. 
In addition, we experiment with taking the whole frames as input of both branches, that is, model 3 and model 6. By comparing model 3 and model 5 (and also model 6 and model 8), we show that the improvement comes not only from architecture of the two-stream block itself but also indeed from the decomposition into structure and detail components. The network with the proposed SD block allows each branch to explicitly focus on reconstructing a single component, which is easier than reconstructing a mixture of multiple components. Each branch makes use of the other one such that it can obtain enough information to reconstruct the high-resolution version for that component. The advantage of the proposed SD blocks can also be observed in the qualitative comparison as shown in Fig.~\ref{fig:model_ablated}.

In addition, we show in Tab.~\ref{table:block} that each model can gain further boost in performance with the proposed HSA module, about 0.04 dB in PSNR and 0.002 in SSIM on average. This module does not only work for the proposed network with SD blocks but also helps improve the performance for the ones with one-stream and two-stream residual blocks. The hidden state adaptation module allows the model to selectively use the history information stored in hidden state, which makes it robust to appearance change and error accumulation to some extent.

\begin{table}[t]
	\centering
	\scalebox{0.8}{	
		\begin{tabular}{c|c|c|c|c}
			\hline
			\multirow1{*}{\textbf{$(\alpha, \beta, \gamma)$}}	
		     &$(1, 0.5, 1)$ &$(0.5, 1, 1)$  & $(1, 1, 0)$  & $(1, 1, 1)$   \\ \hline \hline
			PSNR/SSIM  &27.56/0.8440 &27.77/0.8459 &27.73/0.8453  &27.79/0.8474
			\\	
			\hline
		\end{tabular}
	}
	\vspace{3mm}
	\caption{Ablation study on influence of different loss items.}
	\vspace{-5mm}
	\label{table:weights}
\end{table}
{\flushleft{\textbf{Influence of different components.}}}
The above experiment shows that decomposing the input into two components and processing them with the proposed SD blocks brings much improvement. We also investigate the relative importance of these two components by imposing different levels of supervision on the reconstruction of two components. It implies that the relative supervision strength applied to different components also plays an important role in the super-resolution performance. As shown in Tab.~\ref{table:weights}, when the weights for structure component, detail component and the whole frame are set to $(\alpha, \beta, \gamma) = (1, 1, 1)$, it achieves a good performance of $27.79$/$0.8474$ in PSNR/SSIM. The performance degrades when the weigh for structure component more than the weight for detail component (\ie $(\alpha, \beta, \gamma) = (1, 0.5, 1)$), and verse vise (\ie $(\alpha, \beta, \gamma) = (0.5, 1, 1)$). The result of $(1, 1, 0)$ is $0.06$dB lower than that of $(1, 1, 1)$, which means applying additional supervision on the combined image helps the training of the model. 

\begin{table}[t]
	\centering
	\scalebox{0.625}{	
		\begin{tabular}{l||c|c|c||c|c|c|c|c||c}
			\hline
			Vid4  &\#Frame &FLOPs &\#Param.  &Calendar (Y) &City (Y) &Foliage (Y)  &Walk (Y) &Average (Y)  &Average (RGB)  
			\\
			\hline \hline  	
			Bicubic & 1 &N/A &N/A &18.83/0.4936  &23.84/0.5234	&21.52/0.4438  &23.01/0.7096  & 21.80/0.5426   &20.37/0.5106	
			\\	
			SPMC $^\dagger$~\cite{tao2017detail}  & 3 & - & -   &~~-/-~~~~ &~~-/-~~~~ &~~-/-~~~~ &~~-/-~~~~ &25.52/0.76~~~ &-/-
			\\
			Liu$^\dagger$~\cite{liu2017robust} & 5 &- &- &21.61/-~~~~~~~ &26.29/-~~~~~~~ &24.99/-~~~~~~~ &28.06/-~~~~~~~ &25.23/-~~~~~~~ &-/-
			\\
			TOFlow~\cite{xue2019video} & 7 &0.81T &1.41M &22.29/0.7273  &26.79/0.7446 &25.31/0.7118 &29.02/0.8799 &25.85/0.7659 &24.39/0.7438	
			\\
			DUF-52L~\cite{jo2018deep} & 7  &0.62T &5.82M  &24.17/0.8161 &28.05/0.8235 &26.42/0.7758 &30.91/0.9165	&27.38/0.8329 &25.91/0.8166
			\\
			RBPN~\cite{haris2019recurrent} & 7 &9.30T &12.2M &24.02/0.8088 &27.83/0.8045 &26.21/0.7579 &30.62/0.9111 &27.17/0.8205  &25.65/0.7997 
			\\
			EDVR-L$^\dagger$~\cite{wang2019edvr} & 7 &0.93T &20.6M  &24.05/0.8147 &28.00/0.8122	 &26.34/0.7635  &{\color{blue}31.02}/0.9152  &27.35/0.8264  &25.83/0.8077   	
			\\	
			PFNL$^\dagger$~\cite{yi2019progressive} &7 &0.70T &3.00M  &23.56/0.8232 &28.11/0.8366 &26.42/0.7761 &30.55/0.9103 &27.16/0.8365 &25.67/0.8189 
			\\
			TGA~\cite{isobe2020video} &7 &0.23T &5.87M  &{\color{blue}24.50}/0.8285 &28.50/0.8442 &26.59/0.7795 &30.96/0.9171 &27.63/0.8423 &26.14/0.8258
			\\
			FRVSR 10-128~\cite{sajjadi2018frame} &recurrent (2) &0.14T &5.05M &22.67/0.7844  &27.70/0.8063 &25.83/0.7541   &29.72/0.8971 &26.48/0.8104 &25.01/0.7917
			\\
			RLSP 7-256~\cite{Fuoli-arxiv19-rlsp} &recurrent (3) &0.09T &4.21M  &24.36/0.8235  &28.22/0.8362 &26.66/0.7821 &30.71/0.9134 &27.48/0.8388 &25.69/0.8153
			\\ 	\hline \hline 
			RSDN 5-128 & recurrent (2)  &0.08T &3.83M &24.34/0.8242 &28.73/0.8374  &26.66/0.7842 &30.73/0.9149   &27.61/0.8402 &26.13/0.8238
			\\
			RSDN 7-128 & recurrent (2)  &0.10T &5.01M &24.46/{\color{blue}0.8305}	 &{\color{blue}29.01}/{\color{blue}0.8480}  &{\color{blue}26.78}/{\color{blue}0.7921}  &30.92/{\color{blue}0.9189}   &{\color{blue}27.79}/{\color{blue}0.8474}  &{\color{blue}26.30}/{\color{blue}0.8314}
			\\
			RSDN 9-128 & recurrent (2)  &0.13T &6.19M &{\color{red}24.60}/{\color{red}0.8355}	 & {\color{red}29.20}/{\color{red}0.8527}  &{\color{red}26.84}/{\color{red}0.7931}  &{\color{red}31.04}/{\color{red}0.9210} &{\color{red}27.92}/{\color{red}0.8505}  &{\color{red}26.43}/{\color{red}0.8349}
			\\
			\hline
		\end{tabular}
	}
	\vspace{3mm}
	\caption{Quantitative comparison (PSNR (dB) and SSIM) on \textbf{Vid4} for $4\times$ video super-resolution. {\color{red}Red} text indicates the best and {\color{blue} blue} text indicates the second best performance. Y and RGB indicate the luminance and RGB channels, respectively. FLOPs (MAC) are calculated on an HR image of size 720$\times$480. `$\dagger$' means the values are either taken from paper or calculated using provided models. } 
	\label{vid4_table}
\end{table}

\begin{table}[t]
	\centering
	\scalebox{0.42}
	{	
		\begin{tabular}{l||c|c|c|c|c||c|c||c|c}
			\hline   \hline  	
			\textbf{UDM10} &~~~~~~~\textbf{Bicubic}~~~~~~~  &~~~\textbf{TOFlow}~\cite{xue2019video}~~~ &~~~~\textbf{DUF-52L}~\cite{jo2018deep}~~~~   &~~~~~~\textbf{RBPN}~\cite{haris2019recurrent}~~~~~~ &~~~~\textbf{PFNL}$^\dagger$~\cite{yi2019progressive}~~~~ &\textbf{FRVSR 10-128}~\cite{sajjadi2018frame} &~~\textbf{RLSP 7-256}~\cite{Fuoli-arxiv19-rlsp}~~&~~~~\textbf{RSDN 7-128}~~~~  &~~~~\textbf{RSDN 9-128}~~~~ 
			\\\hline
			FLOPs  [TMAC]   &N/A &2.17 &1.65 &24.81 &1.88  &0.36 &0.24 &0.28 &0.35 
			\\
			Runtime [ms]  &N/A &1693 &1413 &3567 &295  &137 &49 &79 &94
			\\ 
			Average (Y) &28.47/0.8523 &36.26/0.9438 &38.48/0.9605 &38.66/0.9596 &38.74/0.9627 &37.09/0.9522  &38.48/0.9606 &{\color{blue}39.13}/{\color{blue}0.9645} &{\color{red}39.35}/{\color{red}0.9653}
			\\
			Average (RGB)  &27.05/0.8267 &34.46/0.9298  &36.78/0.9514 &36.53/0.9462 &36.78/0.9514 &35.39/0.9403 &36.39/0.9465 &{\color{blue}37.26}/{\color{blue}0.9548} &{\color{red}37.46}/{\color{red}0.9557}
			\\ \hline  \hline
			\textbf{Vimeo-90K-T} &\textbf{Bicubic} &\textbf{TOFlow}~\cite{xue2019video} &\textbf{DUF-52L}~\cite{jo2018deep}  &\textbf{RBPN}~\cite{haris2019recurrent}~~ &\textbf{EDVR-L}$^\dagger$~\cite{wang2019edvr}  &\textbf{FRVSR 10-128}~\cite{sajjadi2018frame} &~~\textbf{RLSP 7-256}~\cite{Fuoli-arxiv19-rlsp} &~~~~\textbf{RSDN 7-128}~~~~  &~~~~\textbf{RSDN 9-128}~~~~
			\\ \hline 
			FLOPs [TMAC]  &N/A &0.27 &0.20 &3.08 &0.30  &0.04 &0.03 &0.03 &0.04
			\\
			Runtime [ms] &N/A &215 &167 &470 &99 &28 &11 &13 &15
			\\ 
			Average (Y)  &31.30/0.8687 &34.62/0.9212   &36.87/0.9447 &37.20/0.9458  & {\color{red} 37.61}/{\color{red}0.9489}  &35.64/0.9319   &36.49/0.9403 & 37.05/0.9454 &{\color{blue}37.23}/{\color{blue}0.9471}
			\\
			Average (RGB) &29.77/0.8490 &32.78/0.9040  &34.96/0.9313 &{\color{blue}35.39}/0.9340 &{\color{red} 35.79}/{\color{red} 0.9374} &33.96/0.9192  & 34.56/0.9274 &35.14/0.9325 &35.32/{\color{blue}0.9344}
			\\ \hline
		\end{tabular}
	}
	\vspace{3mm}
	\caption{Quantitative comparison (PSNR(dB) and SSIM) on \textbf{UDM10} and \textbf{Vimeo-90K-T} for $4\times$ video super-resolution, respectively. Flops and runtimes are calculated on an HR image size of $1280\times 720$ and $448\times 256$ for UDM10 and Vimeo-90K-T, respectively. {\color{red}Red} text indicates the best and {\color{blue} blue} text indicates the second best performance. Y and RGB indicate the luminance and RGB channels, respectively. `$\dagger$' means the values are either taken from paper or calculated using provided models. } 
	\label{SPMC_table}
	\vspace{-5mm}
\end{table}

\subsection{Comparison with State-of-the-arts}
In this section, we compare our methods with several state-of-the-art VSR approaches, including SPMC~\cite{tao2017detail}, TOFlow~\cite{xue2019video}, Liu ~\cite{liu2017robust}, DUF~\cite{haris2019recurrent}, EDVR~\cite{wang2019edvr}, PFNL~\cite{yi2019progressive}, TGA~\cite{isobe2020video}, FRVSR~\cite{sajjadi2018frame} and RLSP~\cite{Fuoli-arxiv19-rlsp}. The first seven methods super-resolve a single reference within a temporal sliding window. Among these methods, SPMC, TOFlow, Liu, RBPN and EDVR need to explicitly estimate the motion between the reference frame and other frames within the window, which requires redundant computation for several frames. DUF, PFNL and TGA skip the motion estimation process and partially ameliorate this issue. The last two methods FRVSR and RLSP super-resolve each frame in a recurrent way and are more efficient. 
We carefully implement most of these methods either on our own or by running the publicly available code, and manage to reproduce the results in their paper. 
The quantitative result of state-of-the-art methods on Vid4 is shown in Tab.~\ref{vid4_table}, where the number is either reported in the original papers or computed with our implementation. 
In addition, we also include the number of parameters and FLOPs for most methods when super-resolution is conducted on an LR image of size $112\times 64$ in Tab.~\ref{vid4_table}. 

\begin{figure}[t]
	\centering
	\includegraphics[width=\textwidth]{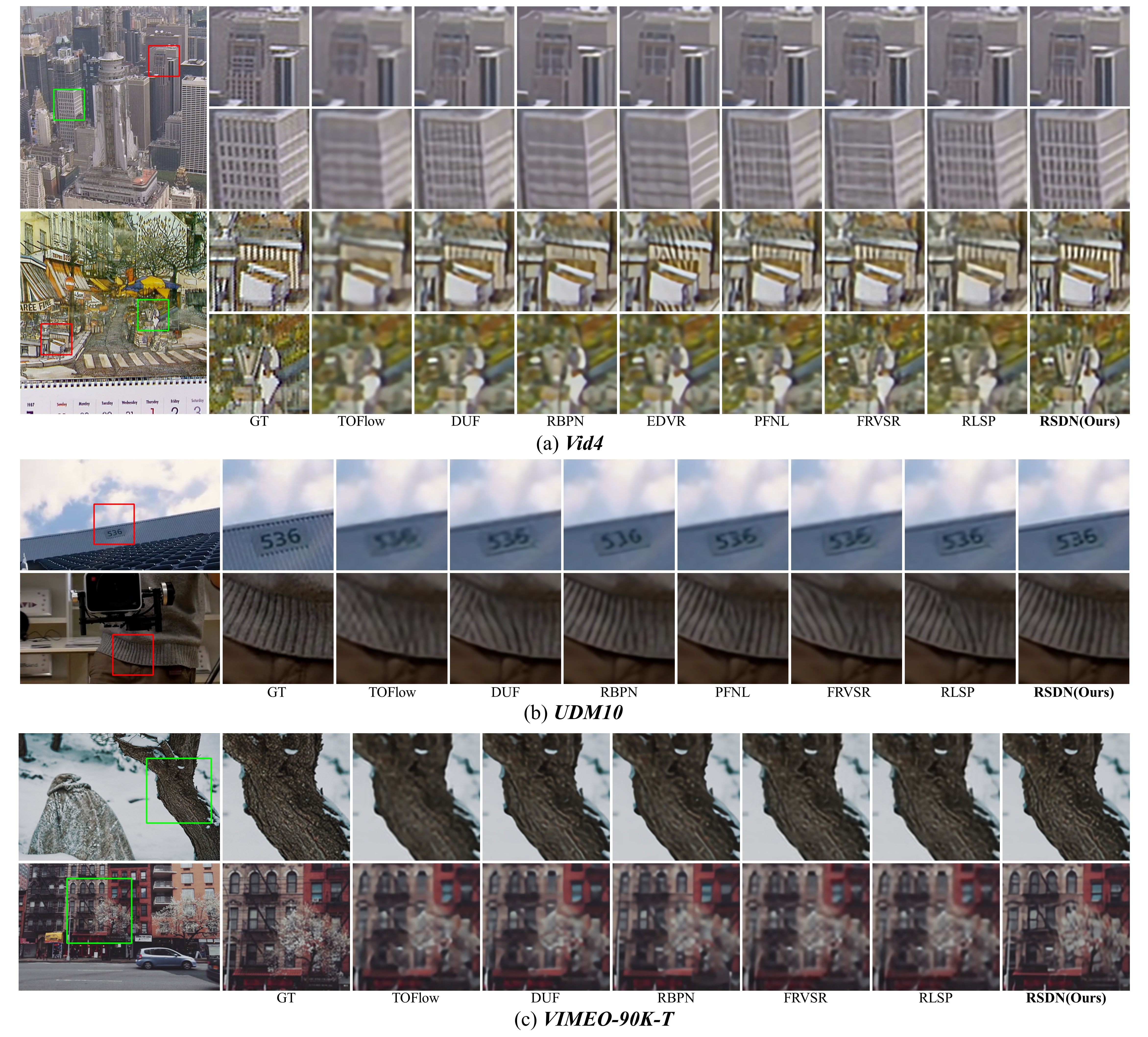}
	\caption{Qualitative comparison on \textbf{Vid4}, \textbf{UDM10} and \textbf{Vimeo-90K-T} test set for $4\times$ SR. Zoom in for better visualization.}
	\label{fig:qualitative}
 	\vspace{-3mm}
\end{figure}
On Vid4, our model with only $5$ SD block achieves $27.61$dB PSNR in Y channel and $26.13$dB PSNR in RGB channels, which already outperforms most of the previous methods by a large margin. By increasing the number of SD block to $7$ and $9$, our methods respectively gain another $0.18$dB and $0.31$dB PSNR in Y channel while with only a little increase in FLOPs. 
We also evaluate our method on other three popular test sets. The quantitative results on UDM10~\cite{yi2019progressive} and Vimeo-90K-T~\cite{xue2019video} two datasets are reported in Tab.~\ref{SPMC_table}. 
Our method achieves a very good balance between reconstruction performance and speed on these datasets.
On UDM10 test set, RSDN 9-128 achieves new state-of-the-art, and is about $15$ and $37$ times faster than DUF and RBPN, respectively. RSDN 9-128 outperforms the recent proposed PFNL, where this dataset is proposed by $0.61$dB in PSNR in Y channel while being $3$ times faster. 
The proposed method is also evaluated on Vimeo-90K-T, which only contains 7-frame in each sequence. In this case, although our method can not take full of its advantage because of the short length of the sequence, it only lags behind the large model EDVR-L but is $6$ times faster.

\begin{figure}[t]
	\centering
	\includegraphics[width=\textwidth]{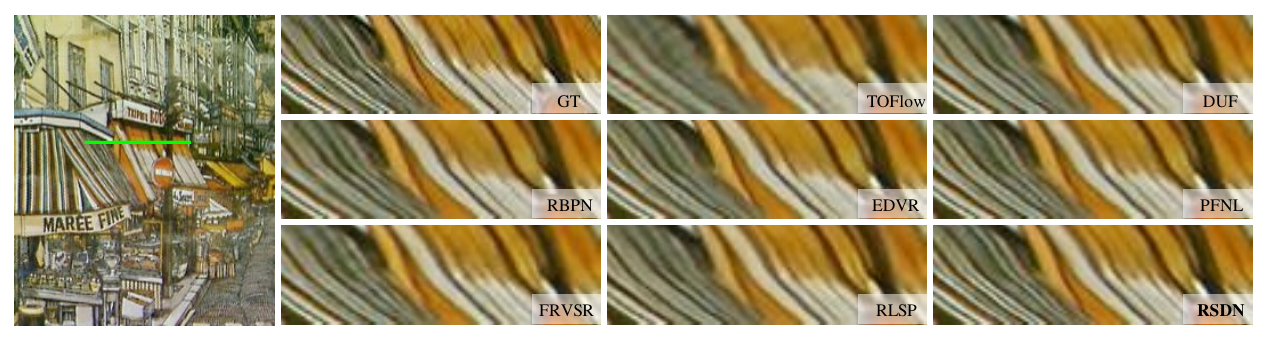}
	\caption{Visualization of temporal profile for the green line on the calendar sequence. 
	}
	\label{fig:profile}
 	\vspace{-3mm}
\end{figure}

We also show the qualitative comparison with other state-of-the-art methods. As shown in Fig.~\ref{fig:qualitative}, our method produces higher quality high-resolution images on all three datasets, including finer details and sharper edges. Other methods are either prone to generate some artifacts (e.g., wrong stripes in an image) or can not recover missing details (e.g., small windows of the building).
We also examine temporal consistency of the video super-resolution results in Fig.~\ref{fig:profile}, which is produced by extracting a horizontal row of pixels at the same position from consecutive frames and stacking them vertically. The temporal profile produced by our method is not only temporally smoother but also much sharper, satisfying both requirements of the video super-resolution task.

\section{Conclusion}
\label{conclusion}
In this work we have presented an effective and efficient recurrent network to super-resolve a video in a streaming manner. The input is decomposed into structure and detail components and fed to two interleaved branches to respectively reconstruct the corresponding components of high-resolution frames. 
Such a strategy allows our method to address different difficulties in the structure and detail components and to enjoy flexible supervision applied to each components for good performance.
In addition we find that hidden state in a recurrent network captures different typical appearance of a scene over time and selectively using information from hidden state can enhance its robustness to appearance change and error accumulation. Extensive experiments on several benchmark datasets demonstrate its superiority in terms of both effectiveness and efficiency.

\bibliographystyle{splncs04}
\bibliography{egbib}
\end{document}